\documentclass[conference]{IEEEtran}
\IEEEoverridecommandlockouts
\usepackage{cite}
\usepackage{amsmath,amssymb,amsfonts}
\usepackage{algorithmic}
\usepackage{graphicx}
\usepackage{textcomp}
\usepackage{xcolor}
\usepackage{psfrag}
\usepackage{mathrsfs}
\usepackage{array}
\usepackage{diagbox}
\usepackage{esint}
\usepackage{bm}
\usepackage{caption}
\usepackage{float} 
\usepackage{subcaption}
\usepackage{cleveref}  
\usepackage{epstopdf}

\def\BibTeX{{\rm B\kern-.05em{\sc i\kern-.025em b}\kern-.08em
		T\kern-.1667em\lower.7ex\hbox{E}\kern-.125emX}}
\begin{document}

\title{Vector Quantized Semantic Communication System 
\\
}
	
\author{Qifan Fu,
		Huiqiang Xie,
		Zhijin Qin,
		Gregory Slabaugh,
		and Xiaoming Tao 
		\thanks{This work was supported in part by the National Natural Science Foundation of China (NSFC Nos 62293484 and 61925105) and in part by Tsinghua University-China Mobile Communications Group Company Ltd. Joint Institute. (Corresponding author: Zhijin Qin.)}
		\thanks{Qifan Fu, Huiqiang Xie and Gregory Slabaugh are with the School of Electronic Engineering and Computer Science, Queen Mary University of London, London E1 4NS, U.K. (e-mail: {q.fu, h.xie, g.slabaugh}@qmul.ac.uk).}
		\thanks{Zhijin Qin and Xiaoming Tao are with the Department of Electronic Engineering, Tsinghua University, Beijing 100084, China (e-mail: qinzhijin, taoxm@tsinghua.edu.cn).
}}

\maketitle
	
\begin{abstract}
		Although analog semantic communication systems have received considerable attention in the literature, there is less work on digital semantic communication systems. In this paper, we develop a deep learning (DL)-enabled  vector quantized (VQ) semantic communication system for image transmission, named VQ-DeepSC. Specifically, we propose a convolutional neural network (CNN)-based transceiver to extract multi-scale semantic features of images and introduce multi-scale semantic embedding spaces to perform semantic feature quantization, rendering the data compatible with digital communication systems. Furthermore, we employ adversarial training to improve the quality of received images by introducing a PatchGAN discriminator. Experimental results demonstrate that the proposed VQ-DeepSC  is more robustness than BPG in digital communication systems and has comparable MS-SSIM performance to the DeepJSCC method.
\end{abstract}
	
	
\section{Introduction}
	As we are moving towards an era of ``Connected Intelligence"\cite{Letaief2019The}, intelligent communication systems have attracted extensive attention. Semantic communication is a new intelligent communication paradigm that has shown many advantages, particularly for the scenarios that involve with large data volume, such as Internet of Things systems and self-driving cars. Compared to the traditional communication system transmitting all the data regardless of its content, semantic communication\cite{qin2021semantic} extracts and transmits the most valuable task-oriented semantic features of the source data and thereby better utilizes communication resources. Existing semantic communication systems can be divided into two categories: data reconstruction  and task execution.
	
	For data reconstruction systems, the transmitter must extract and send the global semantic features of the source data to the receiver, so that the receiver can recover the large and small scale features of the source data. Bourtsoulatze \textit{et al.}  proposed a deep learning (DL) enabled joint source-channel coding (JSCC) system for image transmission, named Deep-JSCC, in which the source-channel encoder and decoder are jointly designed\cite{Bourtsoulatze2019Deep}. 
	Then, Xie \textit{et al.} developed a powerful transformer-based semantic communication system for text transmission, named DeepSC, which is robust to channel variations and applicable to many scenarios\cite{Xie2021Deep}.
	Besides, Weng \textit{et al.} \cite{Weng2021Semantic} designed a speech semantic reconstruction system that leverages an attention mechanism.
	
	To make semantic communication compatible with digital communication systems, semantic features must be converted to bits for transmission. 
	Vector quantization\cite{NIPS2017_7a98af17} can map features into indices, which can then be converted to bits and transmitted by digital communication methods.
	Hu \textit{et al.} \cite{hu2022robust} designed a robust vector quantized semantic communication system to mitigate semantic noise, which outperforms traditional methods on a classification task. Nemati \textit{et al.} \cite{Mahyar2022All} designed a vector quantized DL-based JSCC system and showed its robustness against noisy wireless channels. Although existing vector quantization-based semantic communication systems show good performance for many downstream tasks, their application to data reconstruction remain limited. 
	
	In this paper, we propose a vector quantization-based semantic communication system, named VQ-DeepSC, which uses a semantic embedding space for wireless transmission of images and is compatible with digital communication systems. By sharing semantic embedding spaces between the transmitter and receiver, VQ-DeepSC can perform image transmission with low transmission overheads. The main contributions of this paper can be summarized as follows:
\begin{itemize}
		\item VQ-DeepSC excels at feature extraction and compression to reduce transmitted bits. Specifically, VQ-DeepSC extracts multi-scale semantic features of an image with a U-Net structure \cite{ronneberger2015u}, and compresses these features into transmitted bit stream at transmitter. Then, the image can be reconstructed with multi-scale features at receiver. 
		
		\item  In the model training process, we introduce a PatchGAN discriminator\cite{Isola2017Image} for adversarial training and a generative adversarial network (GAN)-based loss to improve the quality of image generation.
\end{itemize}
	
	Following notations are used in this paper: $\mathbb{R}^{n \times m}$, $\mathbb{Z}^{n \times m}$, and $\mathbb{C}^{n \times m}$ represent sets of real, positive integer, and complex matrices of size $ n \times m $, respectively. Boldface lower-case and upper-case letters denote
	vectors and matrices, respectively. $\odot$ and $ \oslash $ are the element-wise multiplication and division, respectively. ${\left( \cdot \right)^{\rm{H}}}$ denotes the Hermitian. ${\cal C}{\cal N}\left( {\mu ,{\sigma ^2}} \right)$ means the circularly-symmetric complex Gaussian distribution with mean $\mu$ and covariance $\sigma ^2$.

\section{System Model}
\begin{figure} [t]
		\centerline{\includegraphics[width=8cm,height=5cm]{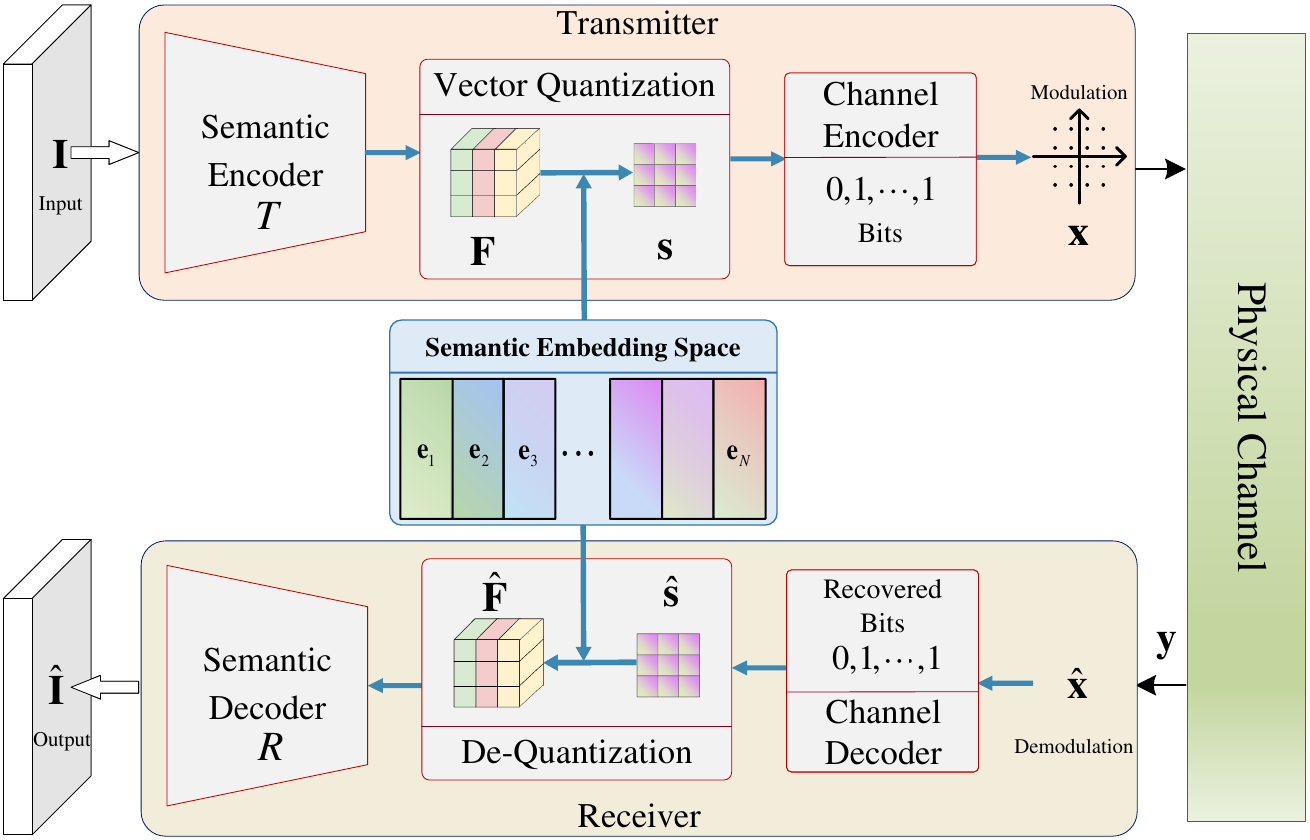}}
		\caption{Vector quantized semantic communication system with multi-scale semantic features.}
		\label{FIG1}
	\end{figure}

	In this section, we introduce the vector quantized semantic communication system with semantic embedding space. As shown in 
	Fig.~\ref{FIG1},
	we consider a DL-enabled end-to-end semantic communication system with a stochastic physical channel, where the transmitter and receiver share a pre-trained semantic embedding space that contains semantic vectors as prior information.
	
\subsection{Transmitter }
	As shown in Fig. 1, the transmitter consists of three parts: semantic encoder, vector quantization with semantic embedding space, and channel encoder.
	We denote the input image as $\bf I$. The semantic feature tensor $\bf F$ is extracted first as 
	\begin{equation} \label{E8}
		\begin{aligned}
			{\bf{F}} = T \left( {\bf I};  {\bm \alpha}  \right),
		\end{aligned}
	\end{equation}
	where $T \left(\cdot ;  {\bm \alpha} \right)$ is the semantic encoder with learnable parameters $ {\bm \alpha} $, and ${\mathbf F}=\left[ {\mathbf f}_1, {\mathbf f}_2, \cdots, {\mathbf f}_M\right] \in {\mathbb R}^{M \times K}$. Specifically, a local semantic embedding space shared by  transceivers is denoted as ${{\bf E} = \left[ {\mathbf e}_1, {\mathbf e}_2, \cdots, {\mathbf e}_N \right] \in \mathbb{R}^{N \times K}}$, where  $N$ represents the number of vectors, and 
	$K$ represents the dimension of each vector. In other words, a semantic embedding space contains $N$ semantic embedding vectors with $K$ dimensions.
	Then with vector quantization\cite{NIPS2017_7a98af17}, the semantic features can be mapped to a set of indices in the embedding space and the transmitter only needs to send these indices rather than the semantic features to the receiver, which performs the nearest neighbor reconstruction:
	\begin{equation} \label{E1}
		\begin{aligned}
			{n} = \mathop {\arg \min }\limits_{n} {\left\| {{\mathbf f}_m - {{{\mathbf e}_n}}} \right\|_2},
		\end{aligned}
	\end{equation}
	where  $ m \in \left[ {1,M} \right], {n} \in \left[ {1,N} \right] $. With the (2), the semantic features can be mapped into the sequence of indices, ${\mathbf s} \in {\mathbb Z}^{M \times 1}$. Then, after conventional channel coding and modulation, ${\mathbf s}$ is processed as a transmitted complex signal $\bm x \in \mathbb{C}^{B \times 1} $, where $B$ is the length of the transmitted signal. 
	
	\subsection{Receiver}
	The receiver also consists of three parts: channel decoder, de-quantization with semantic embedding space, and semantic decoder. The received signal ${\bf y} \in \mathbb{C}^{B \times 1} $ can be expressed as
	\begin{equation} \label{E9}
		\begin{aligned}
			{\mathbf y} = {\mathbf h}  \odot {\mathbf x} + {\mathbf w},
		\end{aligned}
	\end{equation}
	where $\mathbf h$ represents the coefficients of a linear channel between the transmitter and receiver. For the Rayleigh
	fading channel, the channel coefficient follows ${\cal C}{\cal N}(0, 1)$;
	for the Rician fading channel, it follows ${\cal C}{\cal N}\left( {\mu ,{\sigma ^2}} \right)$ with
	$\mu  = \sqrt {{r \mathord{\left/
				{\vphantom {r {\left( {r + 1} \right)}}} \right.
				\kern-\nulldelimiterspace} {\left( {r + 1} \right)}}} $ and 
	$\sigma  = \sqrt {{1 \mathord{\left/
				{\vphantom {1 {\left( {r + 1} \right)}}} \right.
				\kern-\nulldelimiterspace} {\left( {r + 1} \right)}}} $, where $r$ is the Rician coefficient. ${\mathbf w}$ denotes the additive white Gaussian noise (AWGN) whose elements  are i.i.d with zero mean and variance $\sigma^2_n$.  
	
	Assuming that the receiver has perfect channel state information (CSI), the recovered signals are
	\begin{equation} \label{E9}
		\begin{aligned}
			\hat {\mathbf x} = { {\mathbf h}}^{\tt H} \odot {\mathbf y} \oslash \left( {{\mathbf h}} \odot {\hat {\mathbf h}^{\tt H}} \right).
		\end{aligned}
	\end{equation}
	
	Next, the channel decoder recovers the indices $\hat {\mathbf s}$ from ${\bm{\hat x}}$. Then, by selecting the corresponding embedding vectors in the semantic embedding space with $\hat {\mathbf s}$, the receiver can reconstruct the semantic feature tensor $\hat {\bf F}$. Finally, the semantic decoder generates the reconstructed  output ${\bf {\hat I } } = R(\hat {\bf F}; {{\bm \gamma}})$, where $R  \left( \cdot ; {{\bm \gamma}} \right)$ is the semantic decoder with learnable parameters ${{\bm \gamma}}$.

	\section{The Proposed VQ-DeepSC}
\begin{figure*} [t]
		\centerline{\includegraphics[width=16.5cm,height=9.5cm]{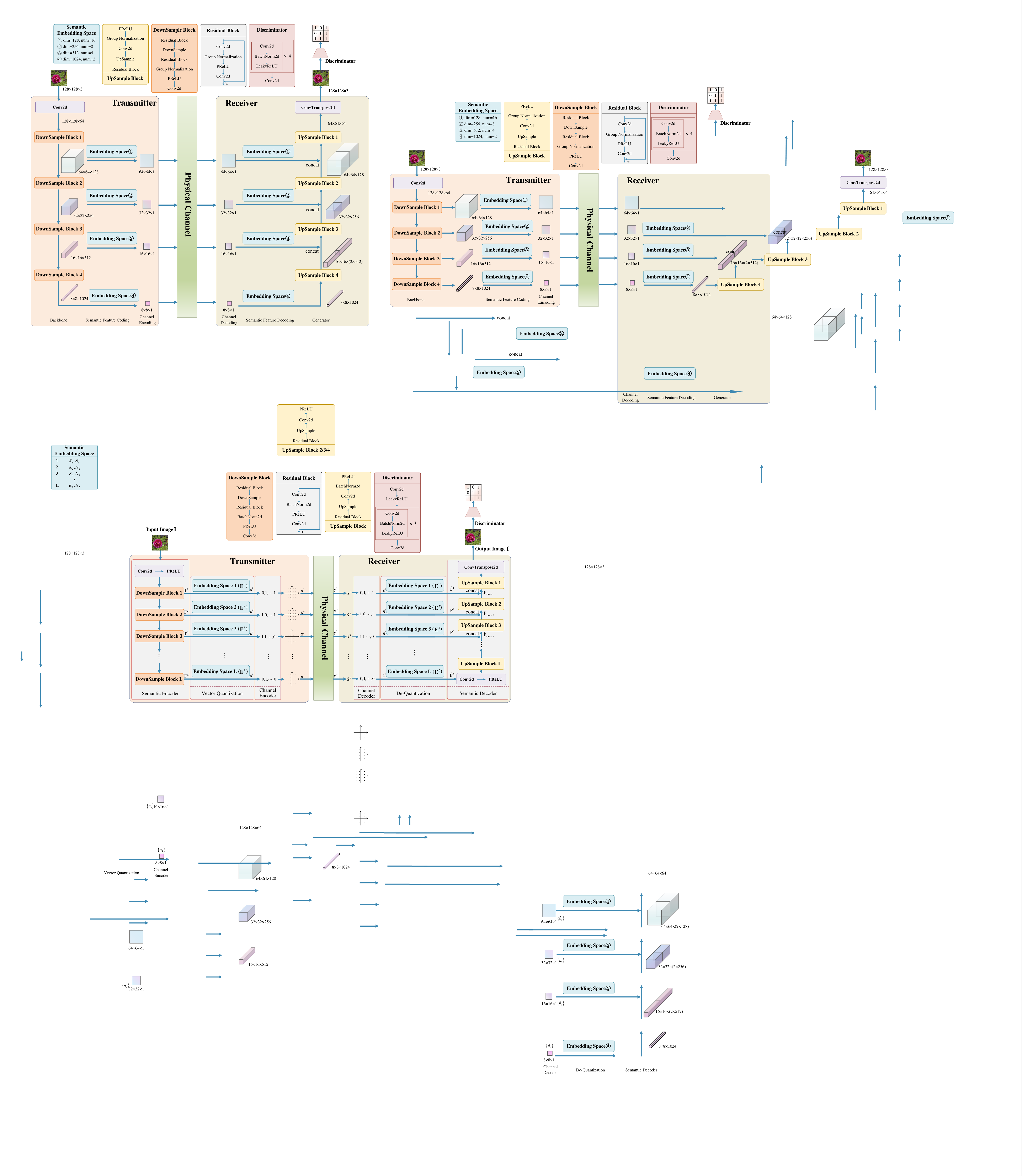}}
		\caption{The multi-scale architecture of VQ-DeepSC represents the embedding space at different scales.}
		\label{FIG2}
	\end{figure*}

	As shown in Fig.~\ref{FIG2}, this section first introduces the framework of VQ-DeepSC, which consists of the transmitter, receiver, and discriminator. Then, the loss function and training of the proposed model are discussed.

	\subsection{Model Description}\label{AAA}
	\subsubsection{Transmitter}
	The transmitter consists of a semantic encoder, vector quantization, and channel encoder. In semantic encoder, a convolution layer with a PReLU activation function and $L$ down-sample blocks are connected sequentially.
	
	The down-sample block shown in Fig.~\ref{FIG2} is successively connected by a residual block, a strided convolutional down-sample layer, and another residual block. After being processed by batch normalization and PReLU activation function, it ends with a convolutional layer as the last layer. The convolutional down-sample layer is a strided convolutional layer, which can reduce the height and width of the feature tensor. The output feature tensor of each down-sample block enters the vector quantization part.
	
	The residual block consists of two convolution layers with a batch normalization and a PReLU activation function between them, and a skip connection between the input of the first convolutional layer and the output of the second convolution layer. Residual blocks can be used to train deeper neural networks to achieve good performance\cite{he2016deep}.
	
	\subsubsection{Receiver} The receiver consists a channel decoder, de-quantization, and semantic decoder. In the semantic decoder, a convolutional layer with a PReLU activation function is used as the input layer, followed by $L$ up-sample blocks and ending with a transposed convolution layer.
	
	The up-sample block consists of a sequence connection of a residual block, an up-sample layer, and a convolutional layer, followed by batch normalization and PReLU activation function. The up-sample layer and the convolutional down-sample layer achieve the opposite effect. 
	In the above layers, only the up-sample layers and the down-sample layers will change the height and width of feature tensor.

	For the pipeline of whole system, the fusion of multi-scale features at the receiver is beneficial to restore detailed features and preserve the structural information of image. 
	Given the input image, $L$ different scales feature tensors ${ {\bf F}^l} $ are extracted by the different down-sample blocks of the semantic encoder,
	\begin{equation} \label{E2}
		\begin{aligned}
			{ {\bf F}^l} = T_{\rm{down}}^l \left( {{{\mathbf F}^{l-1}}};  {\bm \alpha^l}  \right), l = 1, 2, \cdots, L,
		\end{aligned}
	\end{equation}
	where $T_{\rm{down}}^l \left( ;  {\bm \alpha^l} \right)$ represents the $l$-th down-sample blocks in the semantic encoder with learnable parameters $ {\bm \alpha^l} $.
	
	By \eqref{E1}, ${{{\bf F}}^l}$ can be further compressed into ${\mathbf s}^l$ by embedding space ${\mathbf E}^l$. Then, feature indices are converted into bit streams, which are transmitted after channel coding and modulation.

	After transmitting over the physical channel, the received symbols reach the receiver. The receiver inverts the operations performed by the transmitter. The demodulation maps these symbols to bits and the channel decoder decodes the received bits into recovered indices $\hat {\mathbf s}^l$. 
	Next in the de-quantization part, the $\hat {\mathbf F}^l$ is recovered by selecting the corresponding embedding vectors from the embedding space with the $\hat {\mathbf s}^l$. Then, the reconstructed feature tensors of different scales ${\hat {\bf F}^l}$ are fed into the corresponding layers of the semantic decoder. 
	
	In the semantic decoder, feature fusion and reconstruction are carried out through $L$ up-sample blocks. With the recovered $\hat {\mathbf F}^l$, the up-sample block can recover the multi-scale semantic features by
	\begin{equation}
		\hat {\mathbf F}^l_{\text concat} = {\text{concat}}\left(R^l_{up}\left( \hat {\mathbf F}^{l-1}_{\text concat} \right), \hat {\mathbf F}^{l-1} \right),
	\end{equation}
	where ${\hat {{\mathbf F}}^{l-1}_{{\text concat}}} = {{\text concat}}(R^l_{{\rm{up}}}( {{{\hat { {\bf F}}}^l; {\bm \gamma^l}}}), {\hat { {\bf F}}^{l-1}}) $, where ${{\text concat}} \left( \cdot \right)$ represents the concatenation operation. At the end of the receiver, after all up-sample blocks, the image is reconstructed by a transposed convolution layer.

	\begin{figure*}[t]
		\centering
		\begin{subfigure}{0.48\linewidth}
			\centering
			\includegraphics[width=0.9\linewidth]{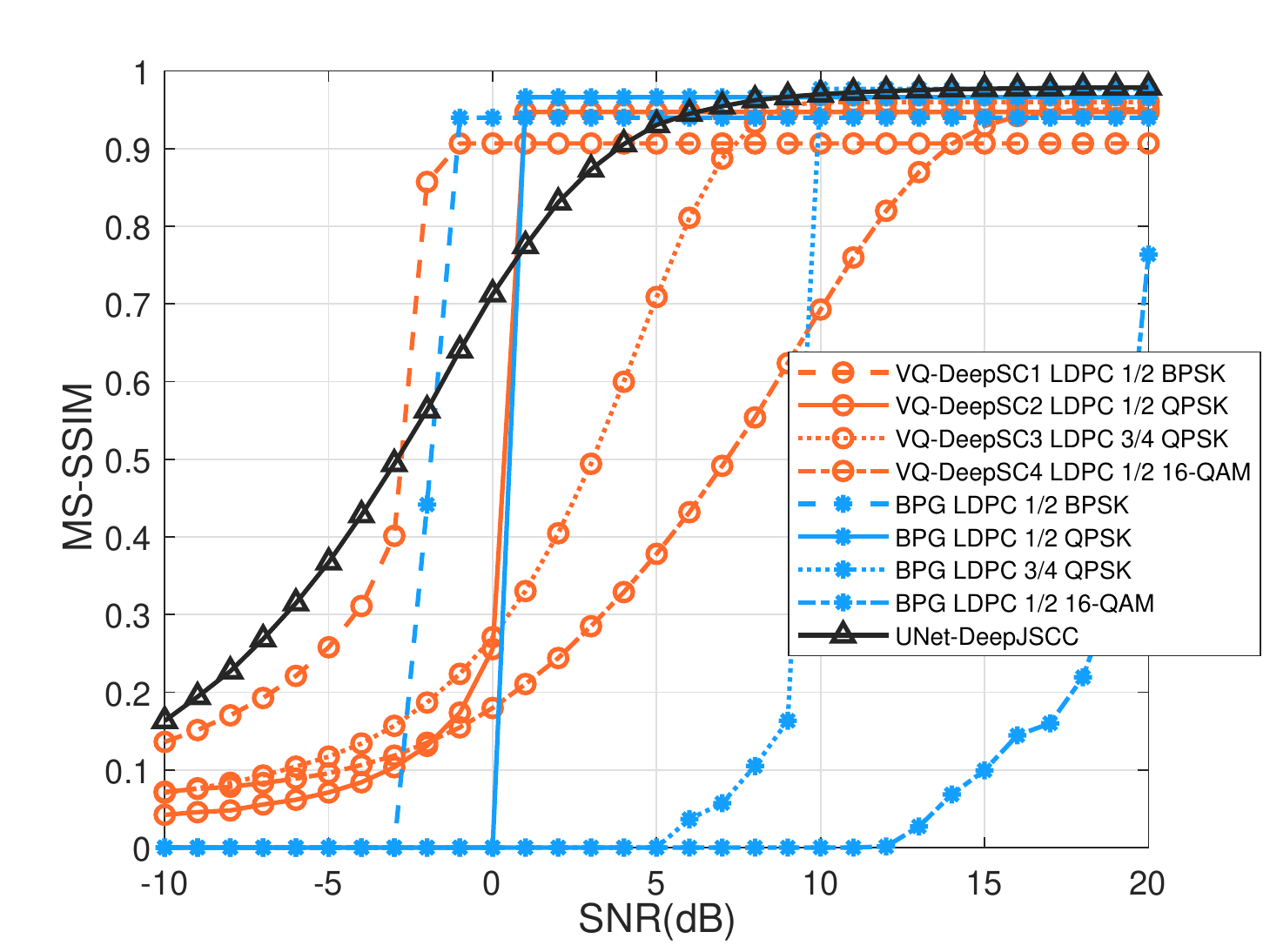}
			\caption{AWGN channels.}
			\label{FIG8}
		\end{subfigure}
		\centering
		\begin{subfigure}{0.48\linewidth}
			\centering
			\includegraphics[width=0.9\linewidth]{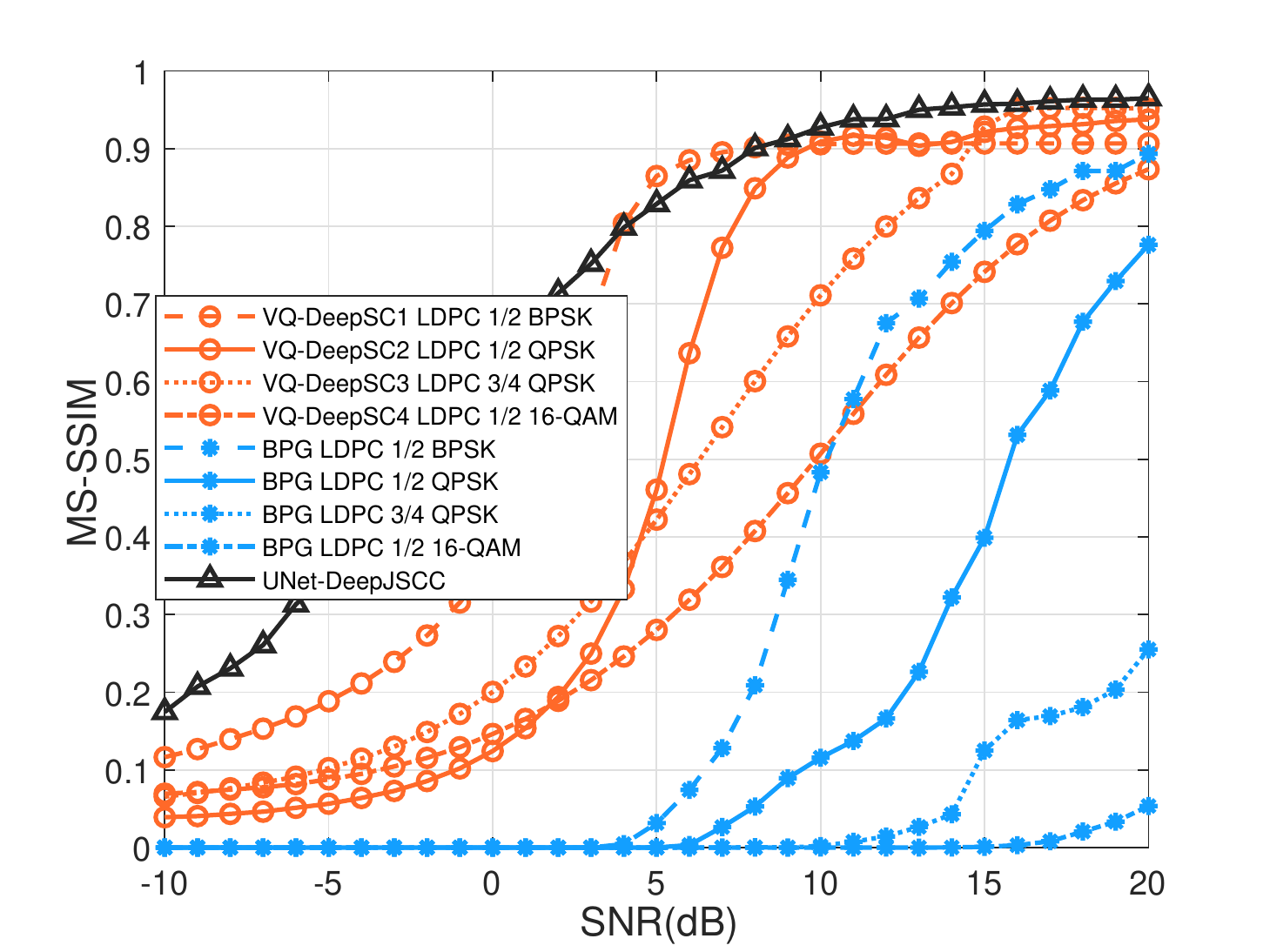}
			\caption{Rician channels.}
			\label{FIG6}
		\end{subfigure}
		\caption{MS-SSIM performance comparisons with a similar number of transmitted  symbols over different channels.}
		\label{SSIM}
	\end{figure*}
	
	\subsection{Loss Function and Training Process}\label{AAA}
	To further improve the quality of the generated image, we introduce a PatchGAN discriminator\cite{Isola2017Image} with binary cross entropy (BCE) loss for adversarial training. The PatchGAN discriminator can make judgements of the patches of the generated image, which makes the model pay more attention to image details. For the input image $\bf I$  that follows distribution ${P_{\bf I}}$, $\hat {\bf I}$  is the recovered image generated by VQ-DeepSC. Then, the loss function of adversarial training is as follows:
	\begin{equation} \label{E3}
		\begin{aligned}
			{{\cal L}_{{\rm{GAN}}}} = \log D\left( {\mathbf I} ; {\mu } \right) + \log ( {1 - D( {\hat {\bf I}}; {\mu } )} ),
		\end{aligned}
	\end{equation}
	where $D\left(\cdot ; {\bm \mu } \right)$ is the discriminator with learnable parameters $\bm \mu $. The introduction of the GAN loss not only enables VQ-DeepSC to generate realistic images, but also enables the model to use unlabeled data for unsupervised training.

	The back-propagation through \eqref{E1} is a non-differentiable due to the quantization operation. To solve this problem, we use a straight-through gradient estimator approach \cite{bengio2013estimating} to copy the gradients from the receiver to the transmitter. In this way, the entire model can be trained in an end-to-end fashion using the following loss function:
	\begin{equation} \label{E4}
		\begin{aligned}
			{{{\cal L}_{{\rm{VQ - DeepSC}}}} =} &{\mathbb E} \left| {{\bf I} - \hat {\bf I}} \right| + \\
			&{\mathbb E} {\left( {\left\| {sg\left[ {\mathbf f}^l_m \right] - {{{{\mathbf e}^l_n}}}} \right\|_2^2 + \beta_c \left\| {sg\left[ {{{{\mathbf e}^l_n}}} \right] - {\mathbf f}^l_m} \right\|_2^2} \right)},
		\end{aligned}
	\end{equation}
	where the first term is the mean absolute error (MAE) loss of the original input image $ {\bf I} $ and generated image $ \hat {\bf I}$, ${{\bf e}_{n}^l} $ is the $n$-th semantic embedding vector from the $l$-th embedding space, ${\mathbf f}^l_m$ is the $m$-th feature vector from the $l$-th feature tensor, $\left\| {sg\left[ {\mathbf f}^l_m \right] - {{\mathbf e}^l_n}} \right\|_2^2$  is the embedding loss of $l$-th layer feature vector and the corresponding embedding space, which moves the embedding vector toward the feature vector in training, $sg\left[ \cdot \right]$ represents the stop-gradient operation, $\left\| {sg\left[ {{{\mathbf e}^l_n}} \right] - {\mathbf f}^l_m} \right\|_2^2$  is the so-called ``commitment loss" \cite{NIPS2017_7a98af17} of the $l$-th layer feature vector and the $l$-th  scale embedding space, which is used to avoid instabilities resulting from large changes in the output of down-sample block, and $\beta_c $  is the commitment loss scalar.

	The final objective of training the entire VQ-DeepSC model to the optimize ${{\cal G}^*} = \left\{ {{ {\bm \alpha}^*},{{\bm \gamma}^*},{{\bm \beta}^*}} \right\}$ as:
	\begin{equation} \label{E5}
		\begin{aligned}
			{{\cal G}^*} = \arg \mathop {\min }\limits_{ {\bm \alpha},{\bm \gamma},{\bm \beta} } \mathop {\max }\limits_{\bm \mu}{\mathbb{E}_{{\bf I} \sim p\left( {\bf I} \right)}}[ \lambda  {{\cal L}_{{\rm{VQ-DeepSC}}}} 
			+ {{\cal L}_{{\rm{GAN}}}} ],
		\end{aligned}
	\end{equation}
	where ${\bm \beta}$ represents the learnable parameters of the embedding spaces, $\lambda $ is a weight to balance  ${{\cal L}_{{\rm{VQ-DeepSC}}}}$ and ${{\cal L}_{{\rm{GAN}}}}$. As \cite{Isola2017Image} shows, the proper combination of the least absolute deviation (LAD) loss and GAN loss can not only improve the clarity of the generated images, but also ensure that the generated images are more realistic.

	\section{Discussions and Simulations}

	This section evaluates the proposed VQ-DeepSC over different channels with perfect channel state information (CSI).
	
	The Cars196 dataset\cite{krause20133d} is used for training 
	and the size of training input image is resized to $256 \times 256$. We save the model with the best performance in training for the final test, which is carried out on the Kodak dataset\footnote{http://r0k.us/graphics/kodak/} and the testing input image size is 768×512.
	Our models were trained in noiseless conditions with the hyper-parameters $\beta_c  = 0.25$ and $\lambda  = 0.1$. 
	Each convolutional layer of semantic encoder and decoder has a $ 3 \times 3 $ kernel.
	Besides, the PatchGAN discriminator\cite{Isola2017Image} begins with a convolutional layer and activation function followed by $3$ sets of convolution layers, batch normalization and activation function, and ends with another convolution layer as output. The size of the first three convolution kernels of the discriminator is $4 \times 4$ and the stride is two, respectively. The size of the last two convolution kernels is $4 \times 4$ and the stride is one.
	The Adam optimizer with learning rate  $ 1.75 \times {10^{ - 4}}$ and StepLR is used for VQ-DeepSC training, with betas of 0.5 and 0.999, batch size of 24, and 400 epochs. The training setting of the discriminator is the Adam optimizer with learning rate  $ 1 \times {10^{ - 5}}$ and LambdaLR, betas of 0.5 and 0.999. The vector dimensions in the four embedding spaces are $K_1 = 128$, $K_2 = 256$, $K_3 = 512$ and $K_4 = 1024$.
	
	In particular, we test the performance of the following methods and evaluated results using the MS-SSIM \cite{wang2003multiscale}
	\begin{itemize}
		\item UNet-DeepJSCC: The remaining 4-layer VQ-DeepSC structures without vector quantization, which transmits images in the form of direct features transmition and compress the transmitted features by $ 1 \times 1 $ convolution. The training channel condations of UNet-DeepJSCC models are SNR = 9 dB for both AWGN and Rician channels.
		
		\item  BPG: The Better Portable Graphics (BPG) \cite{website} codec. 
		
		\item The VQ-DeepSC1: uses three embedding spaces with vector number $N_2=8$, $N_3=4$, $N_4=2$.
		
		\item The VQ-DeepSC2: uses three embedding spaces with vector number $N_2=64$, $N_3=16$, $N_4=4$.
		
		\item The VQ-DeepSC3: uses four embedding spaces with vector number $N_1=2$, $N_2=64$, $N_3=4$, $N_4=4$. 
		
		\item The VQ-DeepSC4: uses four embedding spaces with vector number $N_1=8$, $N_2=4$, $N_3=2$, $N_4=2$.

	\end{itemize}

	Fig. \ref{SSIM} illustrates the MS-SSIM performance comparison for various methods with the similar number of transmitted symbols over AWGN and Rician channels, in which the adaptive modulation and coding (AMC) is employed. Specifically, the LDPC blocklength is 64800 bits.
	In the case of AWGN channels, at the low SNR regimes, the VQ-DeepSC can achieve cliff point about 1 dB lower in SNR than BPG when using the better AMC mode (LDPC with 1/2 rate and BPSK). As the AMC mode changes, the VQ-DeepSC achieves the cliff point at lower SNR regimes than the BPG, which means that the proposed method has better robustness to the channel variations. In addition, VQ-DeepSC can achieve higher MS-SSIM than UNet-DeepJSCC in certain SNR regimes by adjusting coding rates and modulation. For fading channels, the VQ-DeepSC outperforms BPG in terms of MS-SSIM at all SNR regimes and achieve the similar MS-SSIM (above 0.8) as the DeepJSCC when SNR is larger than 5dB. In general, the VQ-DeepSC is more robustness than BPG and has the similar MS-SSIM as the DeepJSCC method.

	\begin{figure} [t]
		\centerline{\includegraphics[width=8cm,height=5.9cm]{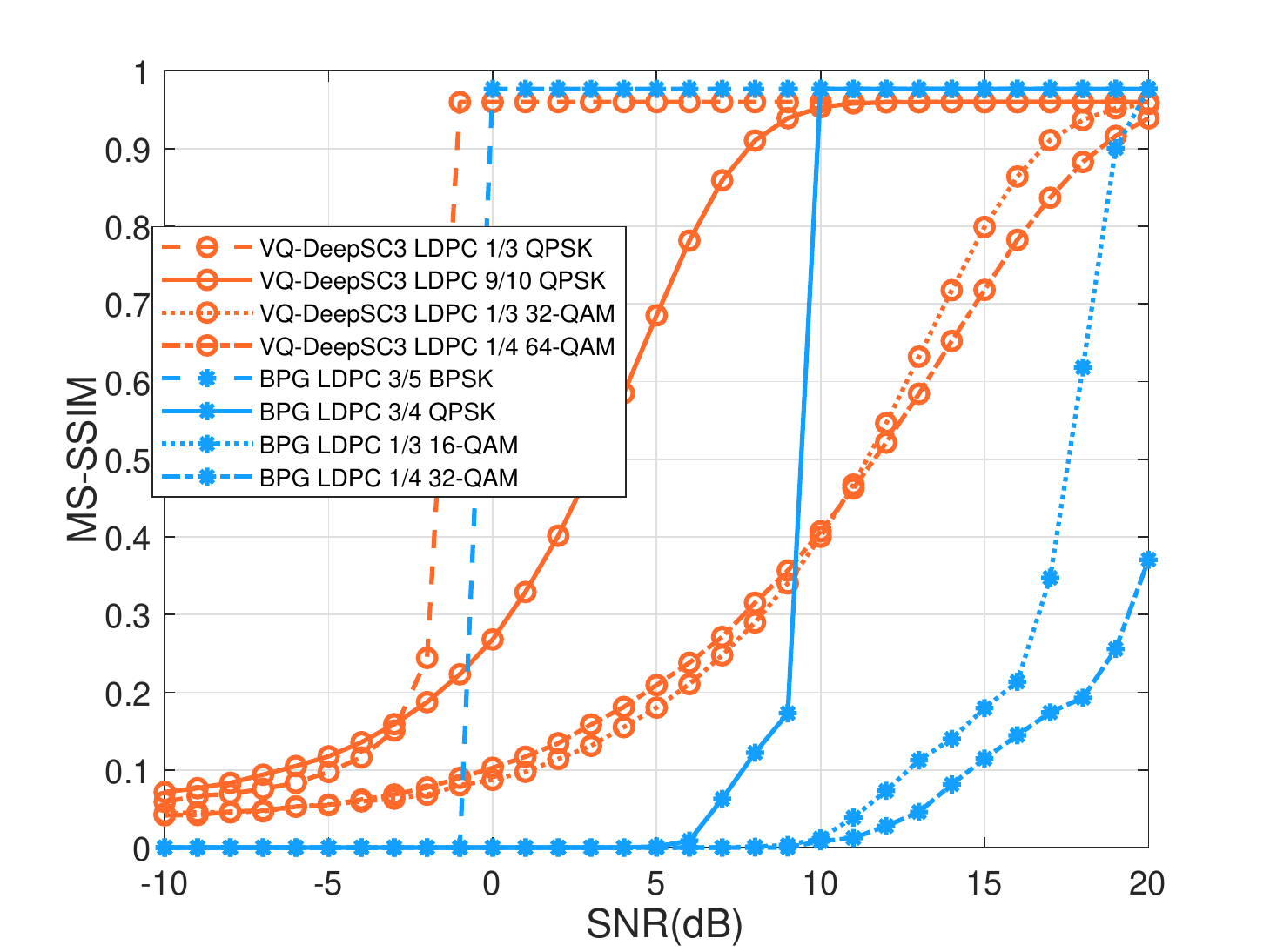}}
		\caption{MS-SSIM performance comparisons with a similar compression ratio of source
			over AWGN channels.}
		\label{2AWGN}
	\end{figure}

	Fig. \ref{2AWGN} shows the MS-SSIM comparison for VQ-DeepSC3 and BPG method with a similar compression ratio of source over AWGN channels. Compared to the BPG, VQ-DeepSC can employ higher modulation order to achieve similar cliff point with higher transmission rate. In Fig. 4, with the same channel coding rate, VQ-DeepSC with 32-QAM has higher MS-SSIM than the BPG with 16-QAM at low SNR regimes, and achieves the same cliff point at SNR=20dB, which shows the high achievable transmission rate and robustness of VQ-DeepSC.

	\begin{figure} [t]
		\centerline{\includegraphics[width=8cm,height=5.9cm]{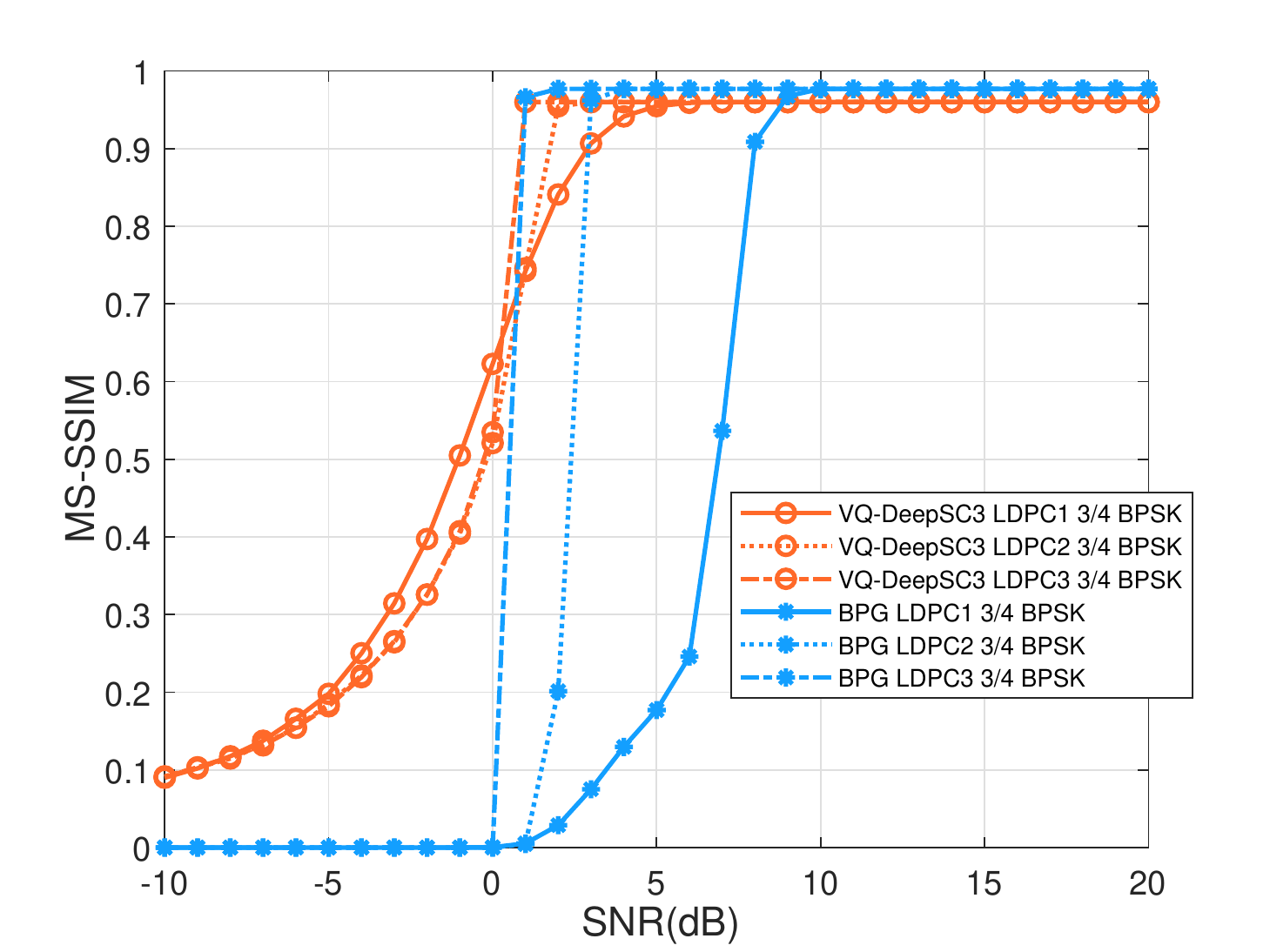}}
		\caption{MS-SSIM comparisons with a similar compression ratio and different LDPC blocklength over AWGN channels.}
		\label{FIG5}
	\end{figure}
	
	Fig. \ref{FIG5} shows the impact of LDPC blocklength on MS-SSIM of VQ-DeepSC3 and BPG with a similar compression ratio of source.
	Specifically, the blocklength of LDPC1, LDPC2, and LDPC3 are 20 bits, 648 bits, and 64800 bits, respectively. BPG with the longer LDPC blocklength outperforms BPG with shorter LDPC blocklength, in which the gain can approximately be 8dB. Obviously, the blocklength of LDPC has less impact on VQ-DeepSC3 than on BPG, which demonstrates the robustness of VQ-DeepSC3 to digital communication systems.

	\section*{V. Conclusions}
	In this paper, we have proposed a framework of vector quantized semantic communication system and introduced a GAN-based implementation for image transmission. Our approach, named VQ-DeepSC, contains different scales of semantic embedding space and realises different scales of semantic features fusion at receiver with low transmission overheads. Specifically, by leveraging vector quantization, the semantic feature vectors were compressed into a set of semantic feature indices in the corresponding embedding space, which also made the model suitable for digital communication systems. In particular, VQ-DeepSC was found to outperform BPG and be comparable to DeepJSCC as measured with MS-SSIM.
	Hence, the proposed VQ-DeepSC is a promising candidate for image semantic communication systems.

	\bibliographystyle{IEEEtran}
	\bibliography{a}

\begin{thebibliography}{10}
\providecommand{\url}[1]{#1}
\csname url@samestyle\endcsname
\providecommand{\newblock}{\relax}
\providecommand{\bibinfo}[2]{#2}
\providecommand{\BIBentrySTDinterwordspacing}{\spaceskip=0pt\relax}
\providecommand{\BIBentryALTinterwordstretchfactor}{4}
\providecommand{\BIBentryALTinterwordspacing}{\spaceskip=\fontdimen2\font plus
\BIBentryALTinterwordstretchfactor\fontdimen3\font minus
  \fontdimen4\font\relax}
\providecommand{\BIBforeignlanguage}[2]{{%
\expandafter\ifx\csname l@#1\endcsname\relax
\typeout{** WARNING: IEEEtran.bst: No hyphenation pattern has been}%
\typeout{** loaded for the language `#1'. Using the pattern for}%
\typeout{** the default language instead.}%
\else
\language=\csname l@#1\endcsname
\fi
#2}}
\providecommand{\BIBdecl}{\relax}
\BIBdecl

\bibitem{Letaief2019The}
K.~B. Letaief, W.~Chen, Y.~Shi, J.~Zhang, and Y.-J.~A. Zhang, ``The roadmap to
  {6G}: {AI} empowered wireless networks,'' \emph{IEEE Commun. Mag.}, vol.~57,
  no.~8, pp. 84--90, Aug. 2019.

\bibitem{qin2021semantic}
Z.~Qin, X.~Tao, J.~Lu, W.~Tong, and G.~Y. Li, ``Semantic communications:
  Principles and challenges,'' \emph{arXiv preprint arXiv:2201.01389v5}, 2021.

\bibitem{Bourtsoulatze2019Deep}
E.~Bourtsoulatze, D.~Burth~Kurka, and D.~Gündüz, ``Deep joint source-channel
  coding for wireless image transmission,'' \emph{IEEE Trans. Cogn. Commun.
  Netw.}, vol.~5, no.~3, pp. 567--579, Sept. 2019.

\bibitem{Xie2021Deep}
H.~Xie, Z.~Qin, G.~Y. Li, and B.-H. Juang, ``Deep learning enabled semantic
  communication systems,'' \emph{IEEE Trans. Signal Processing}, vol.~69, pp.
  2663--2675, Apr. 2021.

\bibitem{Weng2021Semantic}
Z.~Weng and Z.~Qin, ``Semantic communication systems for speech transmission,''
  \emph{IEEE J. Select. Areas Commun.}, vol.~39, no.~8, pp. 2434--2444, Aug.
  2021.

\bibitem{NIPS2017_7a98af17}
A.~van~den Oord, O.~Vinyals, and K.~Kavukcuoglu, ``Neural discrete
  representation learning,'' in \emph{Proc. Adv. Neural Inf. Process. Syst.
  (NIPS)}, vol.~30.\hskip 1em plus 0.5em minus 0.4em\relax Curran Associates,
  Inc., Nov. 2017.

\bibitem{hu2022robust}
Q.~Hu, G.~Zhang, Z.~Qin, Y.~Cai, G.~Yu, and G.~Y. Li, ``Robust semantic
  communications with masked {VQ-VAE} enabled codebook,'' \emph{arXiv preprint
  arXiv:2206.04011}, 2022.

\bibitem{Mahyar2022All}
M.~Nemati and J.~Choi., ``{All-in-One}: {VQ-VAE} for end-to-end joint
  source-channel coding.''
  \emph{DOI:https://doi.org/10.36227/techrxiv.19294622.v1 }, 2022.

\bibitem{ronneberger2015u}
O.~Ronneberger, P.~Fischer, and T.~Brox, ``U-net: Convolutional networks for
  biomedical image segmentation,'' in \emph{Proc. Med. Image Comput.
  Comput.-Assisted Intervention}.\hskip 1em plus 0.5em minus 0.4em\relax
  Springer, Nov. 2015, pp. 234--241.

\bibitem{Isola2017Image}
P.~Isola, J.-Y. Zhu, T.~Zhou, and A.~A. Efros, ``Image-to-image translation
  with conditional adversarial networks,'' in \emph{Proc. IEEE Conf. Comput.
  Vision Pattern Recognit. (CVPR)}, Jul. 2017, pp. 5967--5976.

\bibitem{he2016deep}
K.~He, X.~Zhang, S.~Ren, and J.~Sun, ``Deep residual learning for image
  recognition,'' in \emph{Proc. IEEE Conf. Comput. Vision Pattern Recognit.
  (CVPR)}, Sept. 2016, pp. 770--778.

\bibitem{bengio2013estimating}
Y.~Bengio, N.~L{\'e}onard, and A.~Courville, ``Estimating or propagating
  gradients through stochastic neurons for conditional computation,''
  \emph{arXiv preprint arXiv:1308.3432}, 2013.

\bibitem{krause20133d}
J.~Krause, M.~Stark, J.~Deng, and L.~Fei-Fei, ``3{D} object representations for
  fine-grained categorization,'' in \emph{Proc. IEEE int. conf. comput. vision
  workshops (ICCV)}, Dec. 2013, pp. 554--561.

\bibitem{wang2003multiscale}
Z.~Wang, E.~P. Simoncelli, and A.~C. Bovik, ``Multiscale structural similarity
  for image quality assessment,'' in \emph{Proc. 37th Asilomar Conf. Signals,
  Syst. Comput.}, vol.~2, Nov. 2003, pp. 1398--1402.

\bibitem{website}
F.~Bellard, ``Better portable graphics,'' in \emph{https://bellard.org/bpg/},
  2014.

\end{thebibliography}
\end{document}